\documentclass[letterpaper, 10 pt, conference]{ieeeconf}  

\IEEEoverridecommandlockouts                              

\usepackage{graphics} 
\usepackage{epsfig} 
\usepackage{mathptmx} 
\usepackage{times} 
\usepackage{amsmath} 
\usepackage{amssymb}  
\usepackage{caption} 
\usepackage{multirow}
\usepackage{graphicx}
\usepackage{subfig}
\usepackage{subcaption}
\usepackage[percent]{overpic}
\usepackage{xcolor}
\usepackage{booktabs}
\usepackage{pifont}
\usepackage{hyperref}

\usepackage{arydshln}
\title{\LARGE \bf
 Human Insights Driven Latent Space for Different Driving Perspectives: \\A Unified Encoder for Efficient Multi-Task Inference
}

\author{
Huy-Dung Nguyen$^{1}$,
Anass Bairouk$^{1}$,
Mirjana Maras$^{1}$,
Wei Xiao$^{2}$,
Tsun-Hsuan Wang$^{2}$,
Patrick Chareyre$^{1}$,\\
Ramin Hasani$^{2}$,
Marc Blanchon$^{1}$,
Daniela Rus$^{2}$%
\thanks{$^{1}$Hybrid Intelligence, part of Capgemini Engineering.}%
\thanks{$^{2}$Computer Science and Artificial Intelligence Laboratory (CSAIL), MIT.}%
}

\newcommand{\ie}{\emph{i.e.},~}
\newcommand{\eg}{\emph{e.g.},~}
\newcommand{\ea}{\emph{et~al.}~}

\begin{document}

\maketitle
\thispagestyle{empty}
\pagestyle{empty}
\begin{abstract}
Autonomous driving systems require a comprehensive understanding of the environment, achieved by extracting visual features essential for perception, planning, and control. However, models trained solely on single-task objectives or generic datasets often lack the contextual information needed for robust performance in complex driving scenarios.
In this work, we present a unified encoder trained across a diverse set of computer vision tasks essential for urban driving, including depth estimation, pose estimation, 3D scene flow estimation, and semantic, instance, panoptic, and motion segmentation. This single-encoder approach not only integrates these complementary visual cues, inspired by the diversity of visual cues used in human driving perception, but also enables a compact and inference-efficient model that embeds a rich, navigation-relevant latent space. Indeed, the unified encoder learns to embed multi-task knowledge into a shared representation, allowing for better downstream task adaptation, particularly for steering estimation. To ensure the efficient learning across tasks within a unified encoder, we propose a multi-scale pose decoder and employ knowledge distillation from a multi-backbone teacher model. Our experiments demonstrate that (1) the unified encoder achieves strong generalization across all visual tasks, comparable to state-of-the-art dedicated models, and (2) its frozen latent representations significantly outperform both fine-tuned models and ImageNet-pretrained baselines for steering estimation. These results underscore how multi-task feature learning, inspired by the diversity of perceptual cues used in human driving, offers an efficient and context-rich foundation for autonomous driving systems.
\end{abstract}
\section{Introduction}

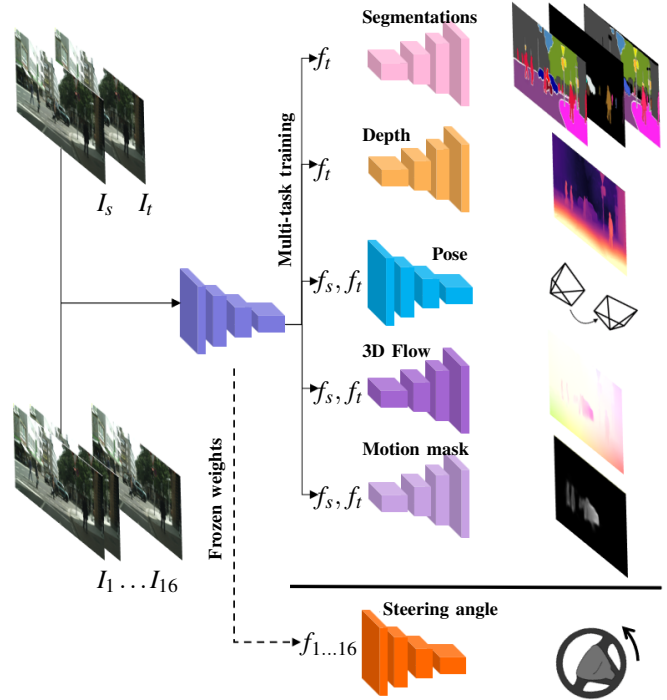
\begin{figure}[!ht]
    \centering
    \begin{overpic}[width=0.95\linewidth]{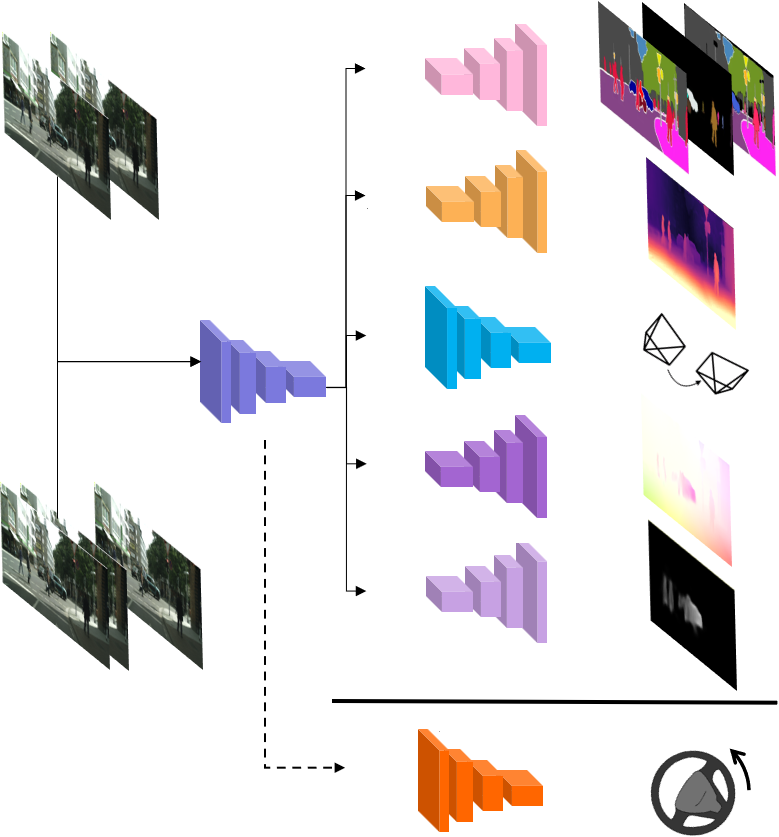}
        \put(12, 70){$I_s$}
        \put(18, 70){$I_t$}
        \put(12, 16){$I_1 \dots I_{16}$}
        \put(43, 91){$f_t$}
        \put(43, 75){$f_t$}
        \put(43, 59){$f_s,f_t$}
        \put(43, 43){$f_s,f_t$}
        \put(43, 28){$f_s,f_t$}
        \put(41, 7){$f_{1 \dots 16}$}
        \put(50, 97){\textbf{\scriptsize Segmentations}}
        \put(50, 80){\textbf{\scriptsize Depth}}
        \put(60, 63){\textbf{\scriptsize Pose}}
        \put(50, 49){\textbf{\scriptsize 3D Flow}}
        \put(50, 35){\textbf{\scriptsize Motion mask}}
        \put(53, 12){\textbf{\scriptsize Steering angle}}
        \put(38, 62){\rotatebox{90}{\textbf{\scriptsize Multi-task training}}}
        \put(28, 20){\rotatebox{90}{\textbf{\scriptsize Frozen weights}}}
    \end{overpic}
    \caption{Our multi-task training strategy. $I_s$, $I_t$, $I_{1 \dots 16}$ represent the source, target, and 16 sequential images, respectively. Their features, denoted as $f_s$, $f_t$, $f_{1 \dots 16}$, are extracted (and concatenated when necessary) using our single encoder.}
    \label{fig:my_model}
\end{figure}

The advancement of self-driving cars has gained significant public attention in recent years, although the development of this technology began decades ago. Early innovations include vehicle-to-vehicle communication via radio waves in the 1920s \cite{milwaukee1926phantom} and electromagnetic guidance in the 1930s \cite{victoria1957power}. The primary goal of autonomous vehicles is to improve road safety and efficiency by minimizing human error, which causes more than 90\% of vehicle accidents, while mechanical failures account for only 2\%~\cite{singh2015critical}. Achieving human-level autonomy requires comprehensive environmental understanding, robust control, and reliable real-time decision making.

A fully autonomous driving system extracts diverse visual cues—such as depth, motion, and segmentation—to support tasks including object detection, path planning, and behavior prediction. Beyond explicit outputs, intermediate visual representations (\eg from CNNs) that capture spatial structure, temporal consistency, and scene dynamics are critical for downstream tasks such as obstacle avoidance, adaptive control, and trajectory prediction \cite{lechner2020neural, kim2017interpretable, xu2016end, bairouk2024exploring}.

Despite their importance, many existing learning-based driving methods solely focus on single-task objectives, such as steering estimation, and are trained using RGB inputs with task-specific supervision \cite{rausch2017learning, sharma2018behavioral, bojarski2016end, do2018real_time}. While these models can learn task-relevant features, their reliance on narrow objectives and input modalities limits the diversity of visual cues captured in the learned representations. Prior work \cite{capito2020optical} has shown that incorporating additional motion information, such as optical flow, can significantly improve steering performance, indicating that richer visual signals can enhance both task accuracy and scene understanding.

In this paper, we propose a novel training strategy for a unified encoder that captures diverse driving-relevant visual cues through multi-task learning. Rather than optimizing isolated objectives, we jointly incorporate tasks that humans implicitly rely on while driving—semantic labeling, object boundaries, depth estimation, and motion cues—into a single perception pipeline (see Figure~\ref{fig:my_model}). This unified design enables compact, inference-efficient multi-task prediction while embedding a rich, navigation-relevant latent representation.

Training a shared encoder across multiple supervised and self-supervised tasks presents significant optimization challenges. In particular, strong supervised segmentation losses can dominate weaker self-supervised objectives such as scene flow and motion mask estimation, destabilizing training and degrading depth accuracy. Addressing these issues is essential to ensure stable convergence and effective feature sharing.

To address these challenges, we introduce a multi-scale pose decoder and apply knowledge distillation from a multi-encoder teacher to balance gradient contributions across tasks. While the resulting training procedure involves multiple stages and an auxiliary teacher model, the additional computational cost is incurred only once during pretraining. Once trained, the unified encoder replaces the multiple task-specific backbones and enables multi-task prediction with a single forward pass, reducing inference-time complexity and deployment overhead. In our setup, the pretraining is conducted on a single NVIDIA A5000 GPU prior to deployment.

Our contributions are summarized as follows:

\begin{itemize}
    \item \textbf{A multi-scale pose decoder}: We design a pose decoder that leverages multi-scale features from the shared encoder, improving depth estimation in dynamic scenes.
    \item \textbf{Knowledge distillation for stability}: We apply knowledge distillation from a multi-encoder teacher model to guide the learning of self-supervised tasks and prevent gradient imbalances.
    \item \textbf{Comprehensive evaluation of shared latent space}: We conduct an evaluation of the unified encoder’s shared representation on a downstream navigation task (steering estimation). Our results demonstrate that the frozen unified encoder outperforms its fine-tuned counterpart and the same architecture pretrained solely on generic datasets such as ImageNet.
    \item \textbf{Human-inspired feature learning}: Our training strategy encourages the encoder to capture richer, human-like perceptual features, promoting contextual awareness and scene understanding closer to how human drivers perceive the surrounding environment.
\end{itemize}

\section{Related work}

\subsection{Image segmentation}
Image segmentation involves separating an image into segments, grouping pixels by specific criteria. Semantic segmentation assigns pixels to broad classes like roads and buildings, while instance segmentation focuses on distinct objects like cars and people \cite{majid2022fast}. Traditionally, these tasks were handled separately. To integrate them, Kirillov \ea \cite{kirillov2019panoptic} introduced panoptic segmentation, which organizes pixels into amorphous background regions ("stuff") and distinct objects ("things"). However, this led to an additional specialized task rather than unifying these two, with performance falling short of state-of-the-art results in dedicated tasks.

Recent methods have proposed unified models for all three tasks showing high performance but still requiring separate training for each. In a significant advancement, Jain \ea introduced OneFormer \cite{jain2022oneformer}, a model that, given an image and a text prompt specifying the task, produces the corresponding type of segmentation. Evaluated on several public datasets, OneFormer set new state-of-the-art benchmarks for all three tasks with a single, jointly trained model. In this paper, we use OneFormer decoders to generate the three mentioned segmentation outcomes.

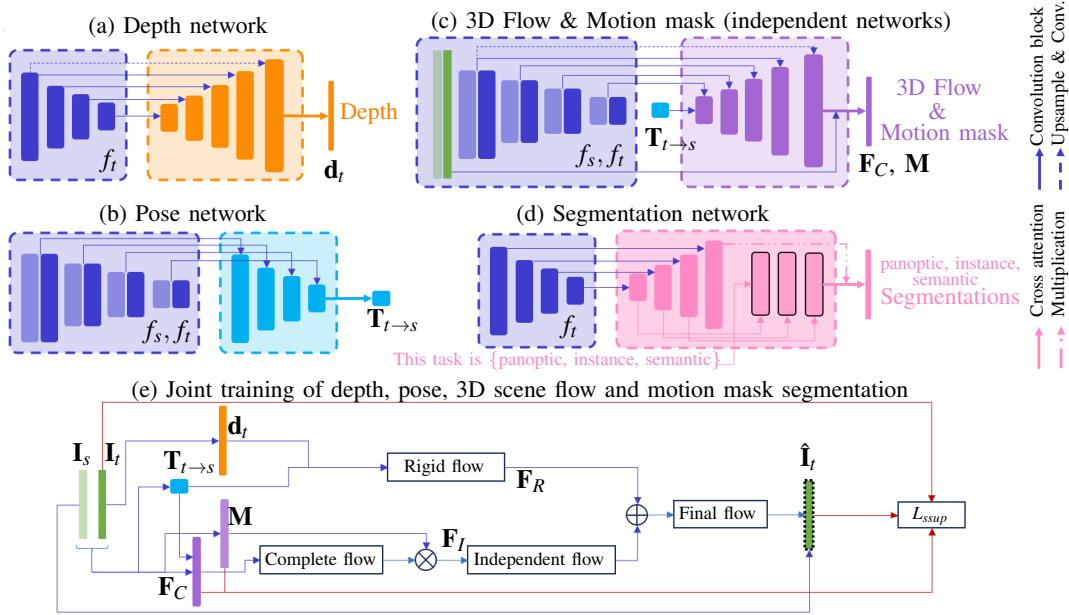
\begin{figure*}[htbp]
    \vspace*{15pt}
    \centering
    \begin{overpic}[width=0.8\linewidth]{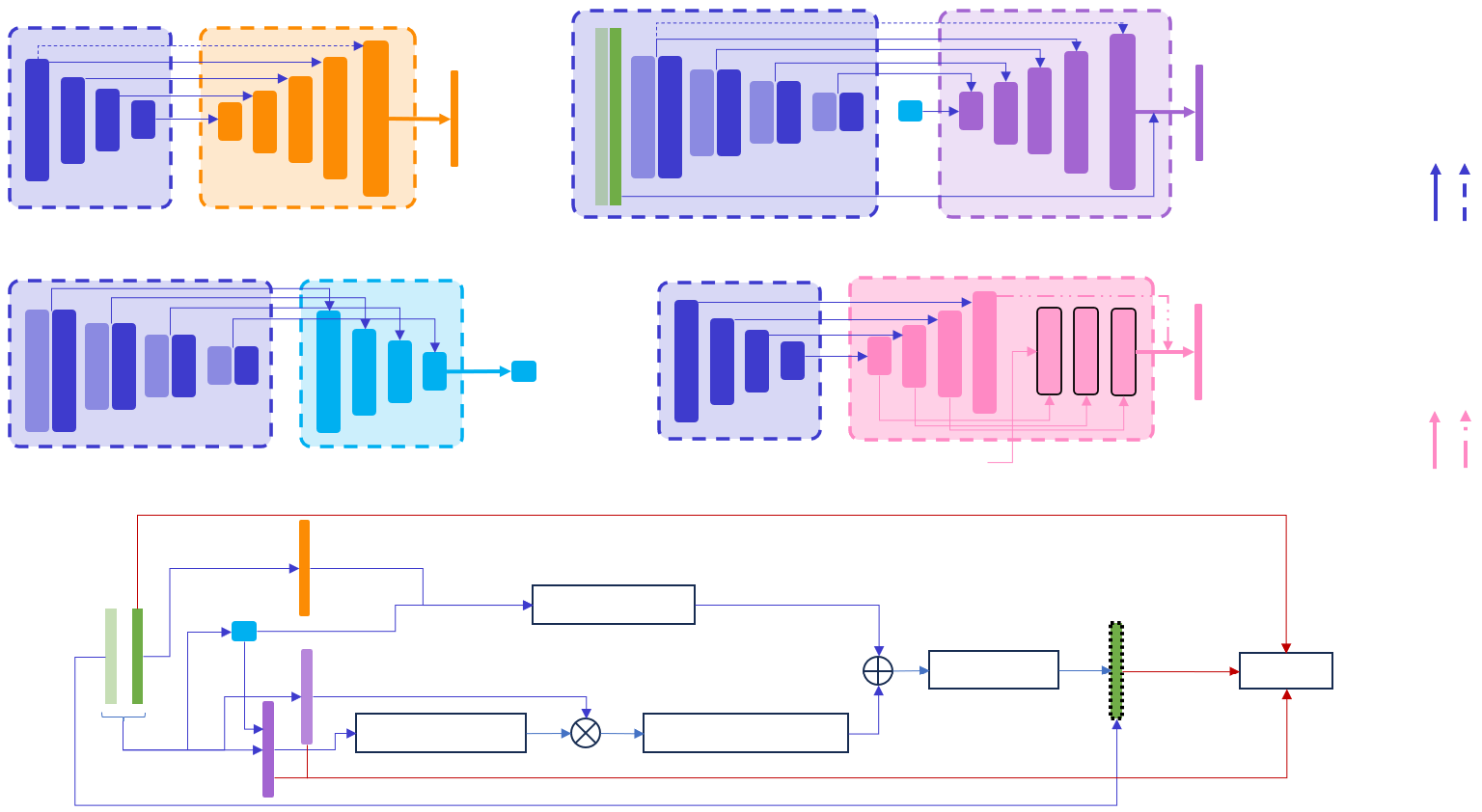}
        \put(9, 42){$f_t$}
        \put(13, 26){$f_s, f_t$}
        \put(53.5, 42.5){$f_s, f_t$}
        \put(52, 26.5){$f_t$}
        \put(31.5, 46.5){\color{orange} \small Depth}
        \put(34, 27){$\textbf{T}_{t \rightarrow s}$}
        \put(60, 44){$\textbf{T}_{t \rightarrow s}$}
        \put(83.5, 48.7){\textcolor[rgb]{0.637, 0.391, 0.820}{\small 3D Flow}}
        \put(86.5, 46.5){\textcolor[rgb]{0.637, 0.391, 0.820}{\small \&}}
        \put(82, 44.5){\textcolor[rgb]{0.637, 0.391, 0.820}{\small Motion mask}}
        \put(82, 32.5){\textcolor[rgb]{1, 0.535, 0.765}{\scriptsize panoptic, instance,}}
        \put(85, 31){\textcolor[rgb]{1, 0.535, 0.765}{\scriptsize semantic}}
        \put(82, 29.2){\textcolor[rgb]{1, 0.535, 0.765}{\small Segmentations}}
        \put(36.5, 23.5){\textcolor[rgb]{1, 0.535, 0.765}{\scriptsize This task is \{panoptic, instance, semantic\}}}
        \put(96, 44){\rotatebox{90}{\scriptsize Convolution block}}
        \put(98, 44){\rotatebox{90}{\scriptsize Upsample \& Conv.}}
        \put(96, 28){\rotatebox{90}{\scriptsize Cross attention}}
        \put(98, 28){\rotatebox{90}{\scriptsize Multiplication}}
        \put(8, 54.5){\small (a) Depth network}
        \put(9, 37){\small (b) Pose network}
        \put(40, 55){\small (c) 3D Flow \& Motion mask (independent networks)}
        \put(48, 37){\small (d) Segmentation network}
        \put(30, 41){$\textbf{d}_t$}
        \put(80, 41.5){$\textbf{F}_C$, \textbf{M}}
        \put(6.5, 14.5){$\textbf{I}_s$}
        \put(9.5, 14.5){$\textbf{I}_t$}
        \put(21, 17){$\textbf{d}_t$}
        \put(15, 13.8){$\textbf{T}_{t \rightarrow s}$}
        \put(14.5, 2){$\textbf{F}_C$}
        \put(21, 8.5){$\textbf{M}$}
        \put(24.5, 4.8){\scriptsize Complete flow}
        \put(44, 4.8){\scriptsize Independent flow}
        \put(41, 6.5){$\textbf{F}_I$}
        \put(37.5, 13.5){\scriptsize Rigid flow}
        \put(48, 12){$\textbf{F}_R$}
        \put(63.2, 9.1){\scriptsize Final flow}
        \put(74.5, 14){$\hat{\textbf{I}}_t$}
        \put(85, 9.1){\scriptsize $L_{ssup}$}
        \put(12, 20.5){\small (e) Joint training of depth, pose, 3D scene flow and motion mask segmentation}
    \end{overpic}
    \caption{Simplified architecture of our model: (a) Depth network using target image features $f_t$ to output depth $\mathbf{d}_t$, (b) Multi-scale pose network using source and target image features $f_s, f_t$ to output relative pose $\mathbf{T}_{t \rightarrow s}$, (c) 3D Scene Flow $\mathbf{F}_C$ and Motion mask $\mathbf{M}$ networks using RGB images and features $f_s, f_t$, (d) Segmentation network outputting panoptic, instance, and semantic segmentations, and (e) Loss computation $L_{ssup}$ for joint training of depth, pose, 3D scene flow, and motion mask segmentation. We denote rigid flow $\mathbf{F}_R$, independent flow $\mathbf{F}_I$, final flow, and sampled target image $\hat{\mathbf{I}}_t$.}

    \label{fig:my_model_2}
\end{figure*}
\subsection{Monocular Depth \& Pose Estimation}
Monocular depth estimation is a task of predicting the depth of a scene from a single 2D image. Unlike stereoscopic methods, which use multiple viewpoints, monocular depth estimation must infer depth from a single viewpoint, making it particularly challenging. Traditional methods mostly based on hand-crafted features, which can lead to inaccuracies in complex scenarios \cite{bhoi2019monocular}. In a first attempt using deep learning, Eigen \ea introduced a CNN-based model to predict depth maps directly from single images, achieving superior results \cite{Eigen2014DepthMP}. However, this method requires ground truth for depth, typically obtained with expensive hardware like LiDAR, limiting its practicality.
To overcome this, recent research has shifted to unsupervised methods that use the inherent structure of unlabeled images. Notably, Zhou \ea \cite{zhou2017unsupervised} and Godard \ea \cite{godard2019digging} proposed unsupervised methods which employ video sequences to simultaneously learn depth and camera motion, reducing the dependency on labeled data and adapting better to dynamic scenes.

However, unsupervised methods often consider the assumption of a static world, which is not true in most real scenarios. To this end, recent advancements have intergrated flow (\eg 2D or 3D) and motion segmentation in addition to depth and pose estimation, enabling better handling of dynamic objects in the scene \cite{sun2023dynamo}.

\subsection{Scene Flow and Motion Segmentation} Scene flow estimation is similar to depth estimation but in addition, it can capture 3D motion between consecutive images, while motion segmentation identifies dynamic objects. These tasks, when jointly trained with depth and pose estimation, improve depth accuracy \cite{hur2020self, hur2021self}. Jiao \ea's EffiScene network \cite{jiao2021effiscene} trains on these tasks using stereo images, leveraging the shared geometric structure of scene depth and object movement. Recently, Sun \ea  \cite{sun2023dynamo} propose DynamoDepth framework that further improves flexibility with the capacity of training on only video frames.

Building on DynamoDepth, we propose a single encoder model that requires only two images to estimate scene flow and motion segmentation, reducing complexity compared to the three-image requirement in the original method.

\subsection{Steering estimation}
The two main research directions for steering estimation are model-based and model-free. Model-based methods rely on vehicle dynamics models, while model-free methods leverage data-driven techniques like deep learning. This paper focuses only on the model-free approach.
Bojarski~\ea \cite{bojarski2016end} introduced an end-to-end steering estimation using CNN with a single RGB camera input. Capito \ea \cite{capito2020optical} improved the navigation capability by incorporating optical flow into the input. The importance of temporal information has also been highlighted in many works. Indeed, Eraqi \ea \cite{eraqi2017end} employed LSTM networks to enhance steering control, while Xu \ea \cite{xu2016end} combined fully convolutional networks with LSTM and semantic segmentation for improved road condition interpretation. Lechner \ea \cite{lechner2020neural} proposed Neural Circuit Policies (NCPs), which provide interpretable decision maps from high-dimensional inputs. These studies indicate that integrating temporal information between frames can improve steering accuracy.

\subsection{Multi-Task Learning in Computer Vision}
Recent works have explored multi-task learning (MTL) frameworks to simultaneously address multiple vision tasks, leveraging shared representations for improved performance. For instance, online knowledge distillation techniques have been proposed to mitigate negative transfer between tasks such as semantic segmentation and depth estimation, enhancing overall learning stability \cite{Jacob2023OnlineKD}. Joint-confidence-guided MTL frameworks have also been introduced for 3D reconstruction, combining depth prediction, semantic labeling, and surface normal estimation to improve feature fusion \cite{wang2023joint}. Moreover, collaborative MTL approaches have demonstrated effectiveness in handling object detection, segmentation, and tracking tasks by establishing associative connections among task heads \cite{Cui2023CollaborativeML}.

Inspired by this synergy, we take a different perspective. Instead of demonstrating how MTL can enhance the performance of one or a few task through training, we hypothesize that the shared features learned from a diverse set of navigation tasks are sufficient not only for achieving good performance on each individual task but also for supporting a related navigation task—steering estimation—which is not present during training.
\section{Methods}

Figure \ref{fig:my_model} illustrates our model architecture, which includes a Swin Tiny encoder \cite{liu2021swin} and six decoders: panoptic, instance, and semantic segmentation, depth and pose, 3D scene flow estimation, motion mask segmentation, and steering command prediction. Training was performed in two stages. In the first stage, we pretrained the encoder with five decoders (excluding the steering command decoder) to learn generalized features across tasks. In the second stage, we froze the encoder and added a prediction head for steering estimation to evaluate its performance in navigation.

Given the challenge of finding an annotated dataset for all the targeted tasks, we selected the CityScapes dataset \cite{cordts2016cityscapes} for training due to its high-quality ground truth segmentation and structured video sequences. This sequential format is particularly suitable for self-supervised training in depth, optical flow, and motion mask estimation without requiring explicit labels. However, we did not evaluate the steering prediction in CityScapes. Instead, this evaluation is performed on the MIT dataset \cite{lechner2020neural} after fine-tuning a prediction head (with the frozen encoder). This MIT dataset provides the appropriate annotations for this task. CityScapes dataset serves as a crucial foundation for training the unified encoder, allowing it to capture rich visual and motion cues that contribute to downstream tasks (\eg steering estimation).

\subsection{Panoptic, Instance, Semantic Segmentations} For these three supervised segmentation tasks, we base on the OneFormer approach. This is a task-conditioned universal image segmentation model that achieves state-of-the-art performance across all segmentation tasks with a unified model and architecture. The OneFormer framework adopts a task-conditioned joint training strategy that allows simultaneous training on all segmentation tasks using a single model, reducing resource requirements significantly.

Figure \hyperref[fig:my_model_2]{\ref{fig:my_model_2}d} provides a simplified illustration of the OneFormer architecture. This model consists of an encoder and two distinct decoders. The first decoder processes image features to generate all segment-related features. The output from this decoder, along with an embedding of a text prompt defining the task (\eg "The task is \{panoptic, instance, semantic\}"), is then passed to the second decoder. The second decoder uses this information to produce the desired segmentation type.
The supervised loss function combines cross-entropy loss for classification, binary cross-entropy for mask predictions, and Dice loss for accurate mask boundary predictions. We trained this model end-to-end on panoptic annotations, from which semantic and instance labels are derived. The overall loss function $\mathbf{L}_{\text{sup}}$ integrates multiple components to ensure task-specific accuracy:
\begin{equation}
\mathbf{L}_{\text{sup}} = \lambda_{\text{cls}} L_{\text{cls}} + \lambda_{\text{bce}} L_{\text{bce}} + \lambda_{\text{dice}} L_{\text{dice}} + \lambda_{\text{contrast}} L_{\text{contrast}},
\end{equation}

where:

\begin{itemize}
    \item $L_{\text{cls}}$ is the cross-entropy loss for mask (\ie region) classification accuracy.
    \item $L_{\text{bce}}$ is the binary cross-entropy loss applied to mask.
    \item $L_{\text{dice}}$ is the Dice loss, helps to improve mask boundaries.
    \item $L_{\text{contrast}}$ is the contrastive loss between object and text queries to make sure the text query is taken into account.
\end{itemize}

\subsection{Depth, Pose, 3D Flow Estimation, Motion Mask}
For depth estimation, we employ the depth decoder architecture from \cite{han2022transdssl}. In addition, we introduce a novel pose decoder, which is discussed in detail in the next section. For 3D scene flow estimation and motion mask segmentation, we build upon the DynamoDepth approach \cite{sun2023dynamo}, enhancing it with convolutional blocks including additional non-linearity to increase the model's representational capacity. Concretely, instead of considering the 3D scene flow as a linear combination of the encoder features $f$ and the pose $\mathbf{T}_{t \rightarrow s}$, we consider it as a non linear combination $\mathbf{F}_{i} = \mathbf{U}(\mathbf{F}_{i+1}) + \textit{ConvELU}([a_i, \textit{Conv}(a_i))]$ where $a_i=\textit{Conv}([\mathbf{U}(\mathbf{F}_{i+1}), f_{i}]])$, $\mathbf{F}_5 = \mathbf{T}_{t \rightarrow s},\text{ and } \mathbf{F} \text{ and } \mathbf{U}$ are respectively flow and upscaling operation. By doing this, we increase the complexity of the decoders, which can help the unified encoder to learn simpler and more generalizable features that can support downstream tasks. Figure \hyperref[fig:my_model_2]{\ref{fig:my_model_2}a}, \hyperref[fig:my_model_2]{\ref{fig:my_model_2}b}, and \hyperref[fig:my_model_2]{\ref{fig:my_model_2}c} provides simplified illustrations of our networks.

Starting with source $\mathbf{I}_s$ and target frames $\mathbf{I}_t$, we estimate depth $\mathbf{d}_t$ and relative pose $\mathbf{T}_{t \rightarrow s}$. Using $\mathbf{T}_{t \rightarrow s}$, we calculate 3D rigid flow $\mathbf{F}_R$. The source and target features are concatenated and passed through the 3D scene flow and motion mask segmentation decoders along with $\mathbf{T}_{t \rightarrow s}$, producing the complete flow $\mathbf{F}_C$ and binary motion mask $\mathbf{M}$. The independent flow $\mathbf{F}_I$ is computed as $\mathbf{F}_I = \mathbf{M} \times (\mathbf{F}_C - \mathbf{F}_R)$. The final flow $\mathbf{F}_F = \mathbf{F}_R + \mathbf{F}_I$ is used to estimate the target frame $\hat{\mathbf{I}}_t$ from $\mathbf{I}_s$. The self-supervision loss $\mathbf{L}_{\text{ssup}}$ is:
\begin{equation}
\mathbf{L}_{\text{ssup}} = L_{\text{recon}} + L_s + \lambda_c L_c + \lambda_m L_m + \lambda_g L_g,
\end{equation}

where:

\begin{itemize}
    \item $L_{\text{recon}}$ is the reconstruction loss that combines SSIM and L1 norms weighted by $\alpha$.
    \item $L_s$ is the smoothness loss \cite{godard2019digging} which is a weighted sum of edge-aware smoothness losses for the inverse depth, the $\mathbf{F}_C$ and $\mathbf{M}$.
    \item $L_c$ is the motion consistency loss that penalize flow discrepancy $\mathbf{F}_D = \| \mathbf{F}_C - \mathbf{F}_R \|_1$ for static pixels.
    \item $L_m$ is the motion sparsity loss that penalizes the mask $\textbf{M}$ with low flow discrepancies using cross-entropy to favor zero motion mask.
    \item $L_g$ is the above-ground loss that penalizes points projected below the ground plane using RANSAC.
\end{itemize}
\begin{figure*}
    \centering
    \includegraphics[width=1.0\linewidth]{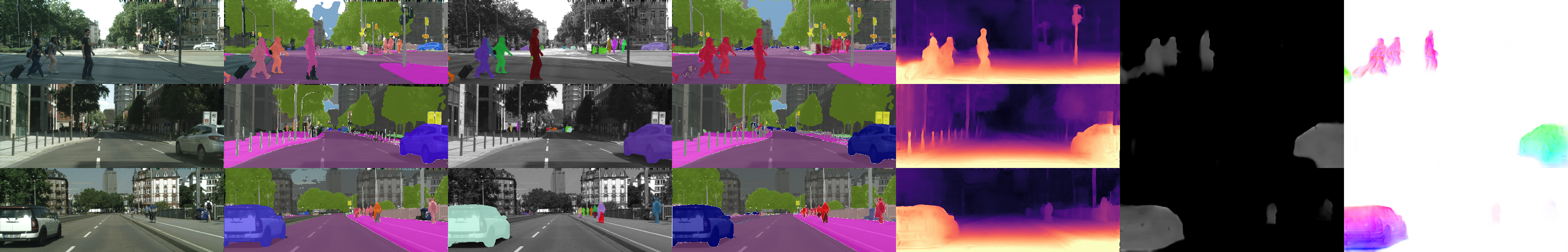}
    \caption{Qualitative results. Left to right: Input, panoptic, instance, semantic output, depth, motion mask, independant flow.}
    \label{fig:qualitative_result}
\end{figure*}

Figure \hyperref[fig:my_model_2]{\ref{fig:my_model_2}e} illustrates the data flow for computing the self-supervised loss described above.

\subsection{Towards a Unified Multi-Task Encoder}

Unlike DynamoDepth, which employs three separate encoders—each specialized for depth estimation, pose estimation, and 3D scene flow with motion mask segmentation—we propose a single, unified encoder capable of handling all these tasks simultaneously, along with semantic, instance, and panoptic segmentation. This unified encoder design is not only for compactness, it is driven by the hypothesis that a unified feature space, with visual and motion cues learned across tasks, can better support both perception and downstream driving-related tasks such as steering estimation. This unified model contrasts with prior systems like OneFormer, which unify only segmentation tasks, or DynamoDepth, which limits multi-task learning to motion and geometry tasks, without extending to semantic scene understanding.
Our proposed multi-task pipeline introduces:
\paragraph{Multi-Scale Pose Decoder}
To address the challenge of low depth estimation accuracy in shared encoder configurations, we introduce a multi-scale pose decoder that progressively refines pose estimates using features at different resolutions. This design may mitigate the ambiguity that arises when pose estimation relies only on low-resolution features, a limitation seen in prior shared-encoder methods like Monodepth2~\cite{godard2019digging}.

The multi-scale pose decoder takes as input the multi-resolution features extracted from the shared encoder for both source $I_s$ and target $I_t$  images. These feature pairs $f$ are concatenated at each resolution level and processed by convolutional blocks with skip connections. The output at each level is computed as:
\begin{equation}
    out = \text{ReLU}\left(\text{ConvBNReLU}\left(\text{ConvBN}(f)\right) + \text{Shortcut}(f)\right),
\end{equation}
Two consecutive blocks are applied per level, with higher-resolution outputs downsampled and concatenated with lower-resolution inputs at each stage. This design preserves both global and local spatial relationships critical for accurate pose estimation, particularly in dynamic environments.
\paragraph{Knowledge Distillation for Stability}
Training a unified encoder across both supervised (\eg segmentation) and self-supervised (\eg scene flow, motion segmentation) tasks presents significant challenges due to the gradient imbalance between these learning types. Supervised tasks, which have strong labeled signals, can dominate the shared feature learning, hindering the convergence of self-supervised tasks like motion mask estimation.

To overcome this, we propose to use knowledge distillation. We first train a teacher model consisting of three encoders and five decoders similar to \cite{sun2023dynamo}: one Swin encoder for depth estimation and the three supervised segmentation tasks, one ResNet18 encoder for pose estimation, and one ResNet18 encoder for 3D scene flow and motion mask segmentation. We then use the encoder-decoder pair responsible for 3D scene flow and motion mask segmentation in the teacher model to supervise these two tasks in the unified encoder model. The associated distillation loss is:
\begin{equation}
L_{distil} = \beta_1 \cdot \|F_C^{teacher} - F_C^{student}\|_1 + \beta_2 \cdot \|M^{teacher} - M^{student}\|_1,
\end{equation}
\textbf{Unified Multi-Task Loss}
The final training objective for the unified encoder integrates supervised, self-supervised, and distillation losses into a single balanced objective:
\begin{equation}
L_{total} = \lambda_1 \cdot L_{sup} + \lambda_2 \cdot L_{ssup} + \lambda_3 \cdot L_{distil},
\end{equation}
\subsection{Steering Estimation from Dense Latent Space}

Steering angle prediction is used here as a proxy task to evaluate whether the learned latent representation captures navigation-relevant structure. While steering alone does not cover the full complexity of driving behavior, it requires implicit understanding of scene geometry, semantics, and motion, making it a suitable first validation target.

For the steering command, we used the pre-trained encoder, which was frozen, followed by an attentive pooling mechanism \cite{chen2022selfattentivepoolingefficientdeep} to process the steering input. The encoder's outputs were fed into the attentive pooler, which takes a sequence of 16 images to compute the steering angle. During training, we optimized the loss $L_{\text{pred}}$, which  is defined as follows:
\begin{equation}
L_{\text{pred}} = {\sum_{i} w_i (\hat{\mathbf{y}}^{(i)} - \mathbf{y}^{(i)})^2} \big/ {\sum_{i} w_i},
\end{equation}
where \(w_i = \exp \left( \lambda \cdot | \mathbf{y}^{(i)} | \right)\), with \(\lambda\) representing the factor modulating the influence of the steering command's magnitude, $|\mathbf{y}^{(i)}|$, on the loss. To evaluate the model's performance, we employed the mean squared error (MSE).

\section{Experimental results}

\subsection{Training Setup} We use the KITTI Eigen split for initial ablation (trained only with the depth and the pose decoders) and CityScapes for the remaining experiments, with image resolutions of $192 \times 640$ and $192 \times 512$, respectively. Images are preprocessed into triples using scripts from \cite{zhou2017unsupervised}. For CityScapes, the lower 25\% of images is cropped to exclude the front car \cite{watson2021temporal}. The entire CityScapes dataset is used for supervised segmentation tasks.
Pretraining is conducted on a single NVIDIA RTX A5000 with a batch size of 6 (3 images for the supervised tasks, \ie segmentations, and 3 triples for the self-supervised tasks) over four steps: (1) training the shared encoder with depth, pose, and segmentation decoders for 60,000 steps, (2) training the 3D scene flow decoder for 40,000 steps with other components frozen, (3) training the pose decoder, 3D scene flow decoder, and motion mask segmentation decoder for 40,000 steps with other components frozen, and (4) training the whole network for 250,000 steps.
After pretraining, we follow the tenfold cross-validation procedure by Lechner \textit{et al.} \cite{lechner2020neural} to fine-tune the model on the downstream steering estimation task. The values of hyperparameters are: $\lambda_{\text{cls}}=2, \lambda_{\text{bce}}=\lambda_{\text{dice}}=\lambda_{\text{c}}=5, \lambda_{\text{contrast}}=0.5, \lambda_{\text{m}}=\lambda_{\text{g}}=\beta_1=0.1, \beta_2=1e-3, \lambda_{\text{1}}=\lambda_{\text{2}}=\lambda_{\text{3}}=1$.
\subsection{Ablation Studies}

\noindent\textbf{Multi-scale pose decoder.} To evaluate the effectiveness of our proposed multi-scale pose decoder, we conducted depth estimation experiments using the KITTI Eigen split dataset. This dataset was selected because accurate depth estimation can be achieved with just a depth and a pose network, allowing us to more clearly isolate and assess the contribution of our multi-scale pose decoder. Table \hyperref[tab:ablations]{\ref{tab:ablations}-1} presents the depth estimation results on the KITTI dataset. Our findings reveal that a naive approach of using a shared encoder, by concatenating the lowest-level features of source and target images and passing them through a pose prediction head, results in a drop in depth estimation performance. This trend is evident in both the Monodepth2 model \cite{godard2019digging} and our Swin encoder. In contrast, using our multi-scale pose decoder maintains depth performance comparable to using a separate ResNet18 encoder specifically for pose estimation.
\begin{table}[ht]
    \centering
    \begin{tabular}{c|ccc|cc|c}
         \multirow{2}{*}{Model} & & & & \multicolumn{2}{c|}{Error (↓)} & Acc. (↑) \\
          & MS & FM & KD & $Abs_{rel}$ & $rmse_{log}$ &  $\delta < 1.25$  \\
    \hline
    $MD2_{sep}$  & & & & 0.115 & 0.193 & 0.877 \\
    
    $MD2_{sh}$  & & & & 0.125 & 0.201 & 0.857 \\
    
    $\text{Our}_{sep}$  & & & & 0.108 & 0.183 & 0.884 \\
    
    $\text{Our}_{sh}$  & & & & 0.117 & 0.190 & 0.872 \\
    
    $\text{Our}_{sh}$  & \ding{51} & & & 0.109 & 0.184 & 0.884 \\
    \hline
    $\text{Our}_{sep}$ & & & & 0.103 & 0.157 & 0.885 \\
    
    $\text{Our}_{sh}$ & \ding{51} &  & & 0.126 & 0.182 & 0.850 \\
    
    $\text{Our}_{sh}$ & \ding{51} & \ding{51} & & 0.134 & 0.183 & 0.833 \\
    
    $\text{Our}_{sh}$ & \ding{51} & \ding{51} & \ding{51} & 0.106 & 0.158 & 0.888 \\
    \end{tabular}
    \caption{\textbf{1)} (Upper part) Ablation on multi-scale pose decoder using KITTI dataset. \textbf{2)}~(Lower part) Analysis on knowledge distilation on CityScapes. MS, FM, and KD refers multi-scale pose decoder, 3D scene Flow \& Motion mask, and Knowledge Distilation. \textit{sep} and \textit{sh} refer to the separate and shared version. For the separate encoder, a ResNet18 was employed similar to \cite{godard2019digging}.}
    \label{tab:ablations}
\end{table}
\begin{table}[hb]
    \centering
    \begin{tabular}{c|c|c|c}
         & $PQ$ (↑) & $AP$ (↑) & $IoU$ (↑) \\
    \hline
    
    OneFormer \cite{jain2022oneformer} & 55.8 & 28.4 & 74.3  \\
    \hline
    Our multi-task model & 56.0 & 28.6 & 74.2 \\
    \end{tabular}
    \caption{Ablation study on panoptic, instance, and semantic segmentation tasks.}
    \label{tab:ablation_segmmentations}
\end{table}

\begin{table*}[t]
\centering
\begin{tabular}{lccccccccccc}
\toprule
\textbf{Method} & \textbf{IM} & \textbf{Sem.} & \textbf{$\#f$} & \textbf{D} & \multicolumn{4}{c}{\textbf{Error metric} ($\downarrow$)} & \multicolumn{3}{c}{\textbf{Accuracy metric} ($\uparrow$)} \\ 
 & & & & & $Abs_{rel}$ & $Sq_{rel}$ & $RMSE$ & $RMSE_{log}$ & $\delta < 1.25$ & $\delta < 1.25^2$ & $\delta < 1.25^3$ \\
\midrule
Monodepth2 \cite{godard2019digging} &  &  & 1 & K & 0.115 & 0.903 & 4.863 & 0.193 & 0.877 & 0.959 & 0.981 \\
LiteMono \cite{zhang2023lite} &  &  & 1 & K & \textbf{0.101} & 0.729 & \textbf{4.454} & \textbf{0.178} & \textbf{0.897} & \textbf{0.965} & \textbf{0.983} \\
Struct2Depth \cite{casser2019depth} & \ding{51} & \ding{51} & 1 & K & 0.141 & 1.026 & 5.290 & 0.215 & 0.816 & 0.945 & 0.979 \\
RM-Depth \cite{hui2022rmdepth} & \ding{51} &  & 1 & K & 0.107 & \textbf{0.687} & 4.476 & 0.181 & 0.883 & 0.964 & 0.984 \\
Dynamo-Depth \cite{sun2023dynamo} & \ding{51} &  & 1 & K & 0.112 & 0.758 & 4.505 & 0.183 & 0.873 & 0.959 & 0.984 \\
\underline{\textbf{Ours} (wo 3D scene flow)} &  & & \underline{1} & \underline{K} & \underline{0.109} & \underline{0.818} & \underline{4.654} & \underline{0.184} & \underline{0.884} & \underline{0.963} & \underline{0.983} \\
\hline
Struct2Depth \cite{casser2019depth} & \ding{51} & \ding{51} & 1 & CS & 0.145 & 1.737 & 7.280 & 0.205 & 0.813 & 0.942 & 0.978 \\
Gordon \ea \cite{gordon2019depth} & \ding{51} & \ding{51} & 1 & CS & 0.127 & 1.330 & 6.960 & 0.195 & 0.830 & 0.947 & 0.981 \\
RM-Depth \cite{hui2022rmdepth} & \ding{51} &  & 1 & CS & 0.100 & \textbf{0.839} & 5.774 & 0.154 & 0.895 & 0.976 & \textbf{0.993} \\
Zhong \ea \cite{zhong2023multi} & \ding{51} &  & 2 & CS & \textbf{0.098} & 0.946 & \textbf{5.553} & \textbf{0.148} & \textbf{0.908} & \textbf{0.977} & 0.992 \\
ManyDepth \cite{watson2021temporal} & \ding{51} & & 2 & CS & 0.114 & 1.193 & 6.223 & 0.170 & 0.875 & 0.967 & 0.989 \\
\underline{\textbf{Ours}} & \underline{\ding{51}} & \underline{\ding{51}} & \underline{1} & \underline{CS} & \underline{0.106} & \underline{1.033} & \underline{5.913} & \underline{0.158} & \underline{0.888} & \underline{0.974} & \underline{0.982} \\

\bottomrule
\end{tabular}
\caption{Depth evaluation on the KITTI (K), and CityScapes (CS) Dataset. \textbf{IM} and \textbf{Sem.} stand for independent motion and semantics. $\#f$ indicates the number of frames during inference. D is the used datatset for training and evaluation. \textbf{Bold} is best. \underline{Underline} is ours.}
\label{tab:sota}
\end{table*}

\noindent\textbf{Knowledge distillation for 3D flow \& motion mask.}
We evaluate the effectiveness of knowledge distillation in training our unified encoder on the CityScapes dataset, which includes many dynamic objects, making 3D scene flow and motion mask networks crucial. Table \hyperref[tab:ablations]{\ref{tab:ablations}-2} shows the depth error and accuracy for different models. First, using a shared encoder with our multi-scale pose decoder achieves good performance compared to using separate encoders for different tasks. Second, training the shared encoder without knowledge distillation leads to suboptimal performance, even worse than the shared encoder model that doesn't consider 3D scene flow and motion masks. However, incorporating knowledge distillation improves the shared encoder's performance, making its performance inline with models with separate encoders. This highlights the effectiveness of our strategy.

\noindent\textbf{Panoptic, instance, and semantic segmentations.} To make a fair comparison, we trained OneFormer using the same batch size as our multi-task model. Table \ref{tab:ablation_segmmentations} shows the results for Panoptic Quality (PQ), Average Precision (AP), and Intersection over Union (IoU) respectively for panoptic, instance, and semantic segmentation tasks. We can observe that the segmentation performance of our multi-task model closely matches that of the OneFormer model, confirming the state-of-the-art capability of our approach.

\subsection{Comparison with State-of-the-Art Methods}

Since our multi-task model aligns closely with the state-of-the-art OneFormer model for segmentation tasks, we only focus on comparing depth estimation with other leading methods. Table \ref{tab:sota} presents depth estimation results on the KITTI and CityScapes datasets.

On KITTI, due to the lack of labeled segmentation data, our model is trained without supervision but still surpasses the average performance of state-of-the-art methods. Notably, it outperforms several established methods (\eg Struct2Depth) across all metrics, even without independent motion and segmentation support. Similarly, on CityScapes, our model shows competitive results, performing better than some methods and approaching the best results for each metric. It is important to highlight that most other methods rely on off-the-shelf segmentation models or use multiple encoders for pose and/or motion estimation.

Finally, Figure \ref{fig:qualitative_result} shows qualitative results of our multi-task model. The panoptic, instance, and semantic segmentation results demonstrate high quality, consistent with the strong quantitative results. Additionally, the quality of the estimated flow and motion mask demonstrate the contribution of these outputs to improved depth estimation compared to the model without the 3D scene flow and motion mask.

\begin{table}[b]
\centering
\resizebox{\columnwidth}{!}{
	\begin{tabular}{lcc}
		\hline
		\textbf{Model}                    & \textbf{Training error} & \textbf{Test error} \\ \hline
		CNN-GRU 64 units \cite{lechner2020neural}                                      & $1.25\pm1.02$                         & $5.06\pm6.64$                    \\
		 CNN-LSTM 64 units \cite{lechner2020neural}                 & $\mathbf{0.19}\pm0.05$                        & $\mathbf{3.17}\pm3.85$          \\
  VAE-LSTM 64 units \cite{bairouk2024exploring}             & $0.54\pm0.26$ & $4.70\pm4.80$            \\ 
  VAE-LSTM 19 units \cite{bairouk2024exploring}             & $0.60\pm0.30$ & $6.75\pm8.33$            \\ \hline
  \underline{(1) Swin-AttnPool (ImageNet pretrained)}           & $2.91\pm2.23$ & $11.03\pm10.03$            \\ 
  \underline{(2) Swin-AttnPool (Encoder unfrozen)}           & $9.62\pm3.24$ & $16.46\pm11.13$            \\ 
  \underline{(3) Swin-AttnPool (Encoder frozen)}           & $1.64\pm1.63$ & $5.41\pm6.06$            \\ 
              
	\end{tabular}
 }

\caption{Results from the passive lane-keeping evaluation across tenfold cross-testing. \textbf{Bold} is best. \underline{Underline} is ours.}

	\label{tab:results}
\end{table}

\subsection{Dense Latent Space to Steering: Evaluation}
Table \ref{tab:results} summarizes the performance of various models for steering prediction, including our proposed Swin-AttnPool approach and several existing CNN-based \cite{lechner2020neural} and VAE-based models \cite{bairouk2024exploring}. To evaluate the effectiveness of the features learned through our training strategy, we present three variants of the same encoder architecture:
\begin{itemize}
    \item ImageNet-pretrained (frozen): The encoder is pretrained on ImageNet, frozen, and fine-tuned on the steering estimation task.
    \item Our pretrained (unfrozen): The encoder is initialized with our pretrained weights and unfrozen during fine-tuning.
    \item Our pretrained (frozen): The encoder is initialized with our pretrained weights and is frozen during fine-tuning.
\end{itemize}

The results reveal several key insights. First, our frozen encoder (variant 3) outperforms the frozen encoder pretrained on ImageNet (variant 1), indicating that the visual features learned through our multi-task training are more relevant for navigation tasks. Second, the frozen encoder (variant 3) also surpasses its unfrozen counterpart (variant 2), suggesting that exclusively fine-tuning for the steering task may overlook valuable features learned from other navigation-related tasks, which can positively contribute to steering performance.

Finally, although Table \ref{tab:results} shows that our training error (variant 3) is relatively high, the test error remains competitive compared to the average performance of all methods. Indeed, our frozen encoder outperforms the VAE-LSTM (19 units) model and performs comparably to the CNN-GRU (64 units) approach. The relatively high error could be due to architectural aspects, either within the encoder or the prediction head, that may not be fully optimized for the steering estimation task.

All experiments are conducted on standard urban driving datasets with predominantly daytime and clear-weather conditions. Evaluating robustness under adverse weather, low-light scenarios, and different geographic domains is an important direction for future work.

\section{Discussion}
Training a unified encoder for multiple tasks introduces significant optimization challenges, particularly in balancing supervised and self-supervised losses. One notable issue was the dominance of segmentation loss over motion-based tasks, resulting in imbalanced feature learning. To address this, we proposed to use knowledge distillation from a multi-encoder teacher, which guided the learning of motion segmentation and scene flow, ensuring stable convergence. Additionally, the proposed multi-scale pose decoder enhanced depth estimation, mitigating inaccuracies in driving scenes. These improvements preserved the advantages of multi-task learning while addressing convergence issues.

Our approach represents an attempt to draw inspiration from human driving perception by integrating visual tasks, such as semantics, geometry, and motion, into a unified representation. We do not explicitly encode cognitive constraints or human priors. Instead, the human-inspired aspect lies in the choice and combination of perception tasks that humans implicitly rely on when driving.

Although the strategy shows promising results, higher numerical errors were observed in certain steering evaluations. These might come from two factors: (1) the used encoder architecture may not be fully optimized for the steering estimation task, and (2) the prediction head was not extensively analyzed, as we only employed an attention pooler among many possible alternatives. Future work should explore other head architectures and refine the model to align more closely with the downstream task, potentially reducing these errors while preserving the benefits of multi-task learning. In addition, future studies can extend the downstream evaluation to other tasks such as trajectory planning, and joint lateral–longitudinal control to explore the generality of the learned representation.

Additionally, while our approach demonstrates the potential of multi-task learning, it introduces more computational complexity due to the simultaneous training of multiple tasks. This additional complexity is primarily incurred during pretraining and is paid only once. In contrast, inference benefits from a single unified encoder instead of multiple task-specific networks. Future work will investigate other distillation strategies and more efficient training to further reduce the computational problem.

Finally, a systematic evaluation of domain generalization, including adverse weather, low-light conditions, and different geographic regions, remains an important direction for future work.
\section{Conclusion}
In this work, we present a unified encoder for autonomous driving, trained using multiple human-inspired visual perception cues necessary for navigation. By integrating knowledge from computer vision tasks—including depth, pose, 3D scene flow estimation, and various segmentation tasks—into a single encoder, our approach enables efficient and compact multi-task inference.

To address the challenges of multi-task training, we introduced a multi-scale pose decoder and applied knowledge distillation from a multi-encoder teacher, ensuring stable convergence and competitive performance for all the pretraining tasks. Our results show that this unified design achieves performance comparable to dedicated models for each task, while enabling coherent steering estimation directly from the shared representation.

Beyond performance, this work highlights the importance of human-inspired perception in autonomous driving. By structuring the encoder to capture rich, contextual information similar to human visual processing, we take a step toward more interpretable and human-centric driving systems. Future work should further explore how human-like perception and reasoning mechanisms can enhance safety and adaptability in complex real-world scenarios.

\bibliographystyle{IEEEtran}
\bibliography{main}

\end{document}